# DIMENSIONALITY REDUCTION AND RECONSTRUCTION USING MIRRORING NEURAL NETWORKS AND OBJECT RECOGNITION BASED ON REDUCED DIMENSION CHARACTERISTIC VECTOR

Name(s) Dasika Ratna Deepthi [1], Sujeet Kuchibholta [2] and K. Eswaran [3]
Address: (1) Ph. D. Student, College of Engineering, Osmania University, Hyderabad-500007, A.P.,
(2) Altech Imaging & Computing, Sri Manu Plaza, A.S. Rao Nagar, Hyderabad, 500062, A.P.,
(3) Sreenidhi Institute of Science and Technology, Yamnampet, Ghatkesar, Hyderabad - 501 301. A.P.,
Country India
Email addresses radeep07@gmail.com, ksujeet@gmail.com, kumar.e@gmail.com

**ABSTRACT**

*In this paper, we present a Mirroring Neural Network architecture to perform non-linear dimensionality reduction and Object Recognition using a reduced low-dimensional characteristic vector. In addition to dimensionality reduction, the network also reconstructs (mirrors) the original high-dimensional input vector from the reduced low-dimensional data. The Mirroring Neural Network architecture has more number of processing elements (adalines) in the outer layers and the least number of elements in the central layer to form "converging-diverging" shape in its configuration. Since this network is able to reconstruct the original image from the output of the innermost layer (which contains all the information about the input pattern), these outputs can be used as object signature to classify patterns. The network is trained to minimize the discrepancy between actual output and the input by back propagating the mean squared error from the output layer to the input layer. After successfully training the network, it can reduce the dimension of input vectors and mirror the patterns fed to it. The Mirroring Neural Network architecture gave very good results on various test patterns.*

# 1. Introduction

This paper proposes a pattern recognition algorithm using a new neural network architecture called Mirroring Neural Network. This paper uses facial patterns as an example, to explain mirroring neural network architecture and illustrate its performance. Facial pattern recognition can be broadly classified into two techniques viz., manually specifying the facial features and automatically extracting the features. This paper deals with the second technique in which neural network recognizes face patterns automatically. There are many problems that can be resolved using neural networks, such as face detection [1] & [2], optical character recognition [3], visual pattern recognition and gender classification [4] etc., Mirroring neural networks are used to mirror the input image pattern and reduce the dimension of the input pattern. This reduced pattern dimension vector (outputs of the central hidden layer) is considered as the signature of the pattern and used for classifying the object. From this reduced pattern dimension vector, the mirroring neural network reconstructs the original image pattern with minimal distortion. The approach is very different from the past work done on recognition and classification of patterns as discussed in [1], [2], [3] and [4] as we used a different architecture coupled with rescaled learning rate parameter and also a different activation function. Mirroring Neural Networks can be used to recognise any type of patterns or object of interest. If many such networks are trained for a multitude of patterns then we have a set of networks that have the ability to recognize patterns albeit one per network. If these networks are connected and a framework or architecture is made such that a pattern is fed as input to all these networks and this architecture gives output from the network which successfully mirrors the pattern, then such an architecture could be a possible data structure for simulated memory. Detailed discussion on this architecture can be found in [5]. In this proposed work we developed the mirroring neural network for face patterns.

# 2. Architecture of Mirroring Neural Network

The present section explains the architecture of the Mirroring Neural Network that can reduce dimension and mirror input patterns, the network can be used to train a single pattern. The Mirroring Neural Network, upon successful training, can recognize the image pattern for which it has been trained. The layered architecture, learning rate parameter and the random weights are configured such that the neural network will converge with a minimal loss of information. The coded form of the object at the least dimensional hidden layer is called signature of the object, it can be used to classify objects. The Mirroring Neural Network contains more adalines in the outer layers and least in the middle layer to form a "converging-diverging" shape as shown in Fig.1.

# Mirroring Neural Network

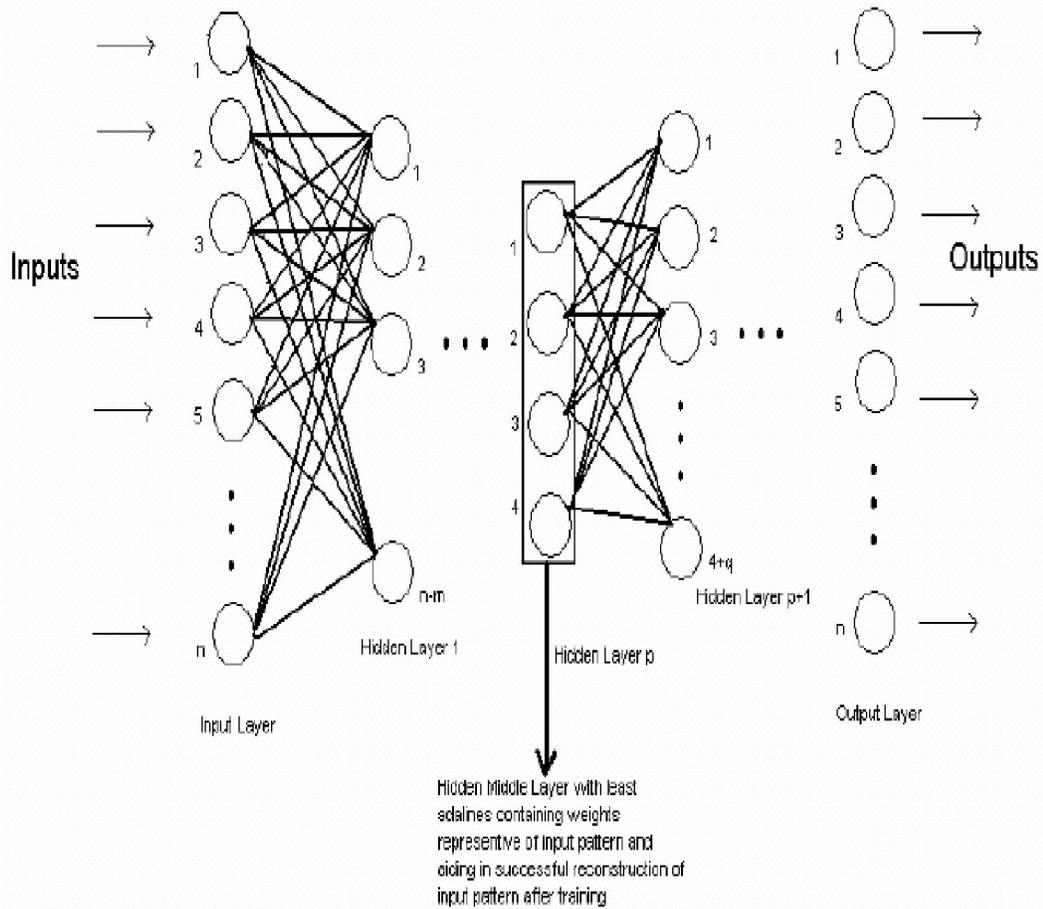

Fig. 1 Typical neural network with inputs, hidden layers and outputs

The converging part of this example network starts with 'n' units at the input layer, '$n_1$' units ($n>n_1$) in the 1st hidden layer, '$n_2$' units ($n_1 > n_2$) in the 2nd hidden layer and so on till it reaches least dimensional hidden layer ($p^{th}$ hidden layer) with 4 units. This converging part condenses the high dimensional pattern into low dimensional code format. The diverging part of the network starts at the least dimensional central hidden layer and ends with the output layer. As we have 4 units at the $p^{th}$ layer of the network, $(p+1)^{th}$ layer will have 4+q units (4+q<n) and so on till it reaches the output layer '$n_L$' having 'n' units (equal to the input vector). The number of hidden layers and the values of variables $n_1$', '$n_2$'... '$n_L$' and q are selected in such a way that the input pattern is mirrored at the output with minimum distortion.

For example, consider a network which has 25 - 10 - 6 - 3 - 8 - 25 nodes in respective layers. This network has 25 inputs, 10 adalines in the 1st hidden layer, 6 adalines in the 2nd hidden layer, 3 adalines in the 3rd, 8 adalines in the 4th and 25 adalines in the last layer. The pattern is reconstructed at the output with its original dimension of 25 units from this signature. The input patterns with 25 dimensions can thus be represented with the 3 code units of the 3rd hidden layer (least dimensional layer). We have tried various architectures with varying hidden layer dimensions. After considerable experimentation, we found that a network having one hidden layer and an output layer is a suitable choice for our pattern. The degree of reduction of the input pattern plays an important role while reconstructing input pattern from reduced dimension vector and so, the number of units in the least dimensional hidden layer must be chosen after careful experimentation. After trying different dimensions of the hidden layers by trail & error method, and checking the neural network's performance, we found that 40 units at the hidden layer gave the most accurate results. We designed our mirroring neural network with 676 inputs to 40 hidden (code) units and 676 output units (676-40-676). The inputs to the network were 26X26 grayscale images.

The input grayscale intensities were rescaled to [0,255]. The rescaled grayscale intensities were then mapped to a range of [-1, +1]. The initial weights were chosen randomly in the range of -0.2 to +0.2. The detailed discussion of this architecture is defined in the following sections.

## 2.1 Input Pattern Rescaling

The input image intensity values were rescaled from 0 to 255 as described in [6].

The rescale function is defined as:

$G_i = (G_i - MIN) * 255 / (MAX - MIN)$
Where
$G_i$ = intensity of the input image
MIN = minimum intensity in the image
MAX = maximum intensity in the image

After rescaling the image pattern, the resulting intensities were mapped to the range of [-1 to +1] using the formula:
$G_i = (G_i - 128) / 128$

Rescaling and mapping was done for every input image presented to the Mirroring Neural Network.

## 2.2 Non-linear Dimensionality Reduction

Initially small random values were chosen for the weights and bias terms of the hidden layer and the output layer. We have used hyperbolic tangent function [8], instead of linear and logistic functions as used in [7], for faster convergence of the network. This differential nonlinear activation function is implemented at each node of the hidden layer and output layer. The transfer function can be defined as:

$$f(x) = \tanh(x/2) = (1 - e^{-x}) / (1 + e^{-x})$$

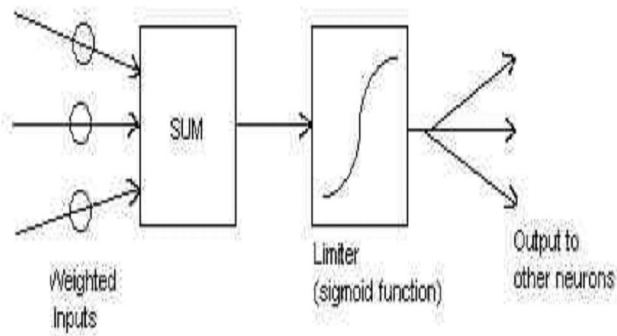

**Fig. 2 Weighted input and use of Sigmoid function to obtain output**

The functionality of any node 'j' in the hidden layer or output layer of the mirroring neural network is shown in Fig. 2 It can be mathematically defined[*] as:

For hidden layer node:

$$NetInput_{hj} = bias_{hj} + \sum_{k=0}^{k=image\ size} W_{hjk} * G_k$$

$$Adaline_{hj} = \tanh(NetInput_{hj}/2)$$

Where  $NetInput_{hj}$ = net input to the $j^{th}$ node in the hidden layer
$bias_{hj}$ = bias term of the $j^{th}$ node in the hidden layer
$W_{hjk}$ = $k^{th}$ weight of $j^{th}$ node in the hidden layer
$G_k$ = $k^{th}$ intensity of the input image
$Adaline_{hj}$ = output of $j^{th}$ node in the hidden layer

For output layer node:

$$NetInput_{oj} = bias_{oj} + \sum_{k=0}^{k=hidden\ layer\ size} W_{ojk} * Adaline_{hk}$$

$$Adaline_{oj} = \tanh(NetInput_{oj}/2)$$

Where  $NetInput_{oj}$ = net input to the $j^{th}$ node in the output layer
$bias_{oj}$ = bias term of the $j^{th}$ node in the output layer
$W_{ojk}$ = $k^{th}$ weight of $j^{th}$ node in the output layer
$Adaline_{hk}$ = output of $k^{th}$ node in the hidden layer
$Adaline_{oj}$ = output of $j^{th}$ node in the output layer

While training the back propagating Mirroring Neural Network we have used the usual gradient descent [10] to minimize the mean squared error between the input and its reconstruction at the output. The activation function and variable learning rate parameter [11] reduce out-of-range values and help in faster convergence of the network. The learning rate parameter was incremented by 10% at the hidden layer compared to the output layer. The mirroring neural network, with learning rate rescaling in combination with the hyperbolic tangent function, learnt the input patterns rapidly and reconstructed them with low deformation.

---

[*] This mathematical definition is pertinent to the neural architecture specified in this paper.

## 2.3 Object Recognition

After successfully training the neural network to a desired accuracy, the average feature vector for the input pattern was computed by averaging the least dimensional hidden layer outputs for all input images. The recognition of the object was based on two threshold values. The first threshold value was the Euclidian distance[†] between the reduced dimension feature vector for the test image and the average feature vector which was computed after the training of the Mirroring Neural Network. The second threshold value was the Euclidian distance between the output and the input of the Mirroring Neural Network. These two values were computed for each test image and if they were within an accepted threshold then we categorized the test image as a face pattern. The threshold values were fixed after considerable experimentation in order to maximize success rate and reduce false acceptance.

## 2.4 Results and Discussion

A training set consisting of 549 facial images (8 Bit) of size 26 X 26 was fed to the mirroring neural network. After sufficient training (success rate above 95%), the network could recognize faces and reject non-face patterns. Recognition was based on the two threshold values discussed in 2.3. Typical sample input images from the training set and their corresponding mirror images (outputs of the mirroring neural network) are shown in Fig. 3. Test samples of face and non- face images are given in Fig. 4.

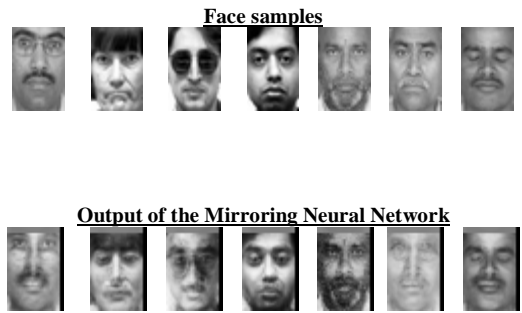

**Fig. 3 Typical face samples and their mirror images**

† The Euclidian distance is computed after normalization.

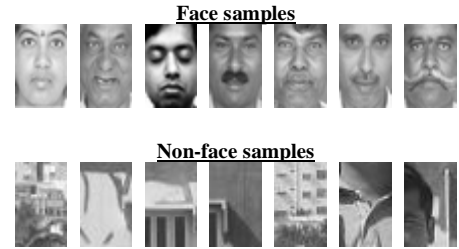

**Fig. 4 Typical face & non-face images**

We tested the network with 250 face images of which the algorithm classified 245 face images correctly for an accuracy of 98%. We also tested 300 more images containing 150 face and 150 non-face images. The algorithm could correctly classify 277 images out of 300 (92.33%). Both sets of test images (250 + 300 = 550) were entirely new images, none of which were in the training set.

We also used training images of size 26X42 to train the network. We increased the nodes in the hidden layer to 70 and the network performed almost equally well.

## 3. Conclusions and future work

The architecture described in this paper is a simple approach for object recognition which is applicable to various image categories like faces, furniture, flowers, trees, etc and was tested for the same with slight changes in the network architecture w. r. t., hidden layer size and threshold values. Such networks could be used for face detection by incorporating them in an application to verify possible face candidates. Such networks can be "parallelized" for recognition tasks pertaining to different patterns. For example, the input may be an image of any pattern which is sent to all the specialized mirroring neural networks in parallel and the network which gives the least value below the two thresholds would identify the input pattern. The overview of this multiple pattern mirroring architecture is illustrated in Fig. 5.

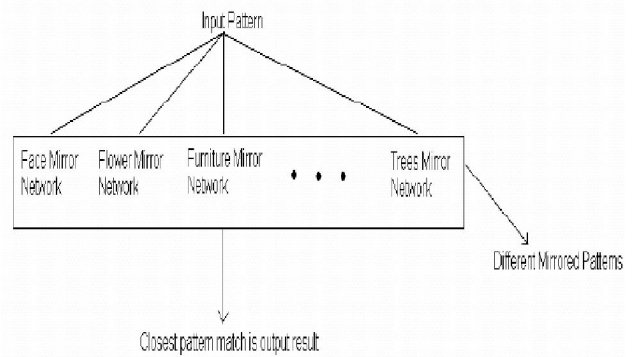

**Fig. 5 Pattern input to multiple mirror networks in parallel**